# Cascaded Asymmetric Local Pattern: A Novel Descriptor for Unconstrained Facial Image Recognition and Retrieval

[1]Soumendu Chakraborty[*], [2]Satish Kumar Singh, and [2]Pavan Chakraborty

[1]Indian Institute of Information Technology, Lucknow, India
[2]Indian Institute of Information Technology, Allahabad, India

[*]Corresponding Author Mobile: +919897043787

[*]soum.uit@gmail.com, sk.singh@iiita.ac.in, pavan@iiita.ac.in

**ABSTRACT**

Feature description is one of the most frequently studied areas in the expert systems and machine learning. Effective encoding of the images is an essential requirement for accurate matching. These encoding schemes play a significant role in recognition and retrieval systems. Facial recognition systems should be effective enough to accurately recognize individuals under intrinsic and extrinsic variations of the system. The templates or descriptors used in these systems encode spatial relationships of the pixels in the local neighbourhood of an image. Features encoded using these hand crafted descriptors should be robust against variations such as; illumination, background, poses, and expressions. In this paper a novel hand crafted cascaded asymmetric local pattern (CALP) is proposed for retrieval and recognition facial image. The proposed descriptor uniquely encodes relationship amongst the neighbouring pixels in horizontal and vertical directions. The proposed encoding scheme has optimum feature length and shows significant improvement in accuracy under environmental and physiological changes in a facial image. State of the art hand crafted descriptors namely; LBP, LDGP, CSLBP, SLBP and CSLTP are compared with the proposed descriptor on most challenging datasets namely; Caltech-face, LFW, and CASIA-face-v5. Result analysis shows that, the proposed descriptor outperforms state of the art under uncontrolled variations in expressions, background, pose and illumination.

## 1. Introduction

In recent years hand crafted descriptors have gained much of the attention in face recognition and facial image retrieval problems. Many hand crafted descriptors have been proposed to improve the retrieval and recognition accuracies under varying environmental as well as physiological traits. A descriptor characterizes the facial image in such a way that, it decreases the intra class whereas increases the inter class dissimilarity [34]. Fisher-face [1], Eigen-face [1], different versions of PCA [2-6], and Linear Discriminant Analysis (LDA) [7-9] were proposed to capture the feature points for face recognition under controlled environment. In recently published literatures, deep learning based image descriptors have been proposed, where a Convolutional Neural Network (CNN) is trained to classify the facial images [10-11] [31]. The accuracy of the CNN based model depends upon the size of the training dataset. To get reasonable accuracy, it requires bigger and bigger dataset. Moreover the machine learning based methods are biased to training data type and may not be generalized for many applications [29][30]. The proposed descriptor is a hand-crafted descriptor and training and testing of these descriptors are not required. VGG model for face recognition is different in its characteristics from hand crafted descriptors, as features are extracted using Convolutional Neural Networks (CNNs). Age invariant Cross-Age Reference Coding (CARC) [32] has been proposed to encode the low level facial features in the age invariant reference space. Most recently Hierarchical Bayesian Model for facial expression recognition under pose variation has been proposed in [33]. Local Binary Pattern (LBP) [12] is one of the most popular descriptors defined in the local neighbourhood of a facial image. Initially texture classification problem was addressed using LBP and later on it was extended for face recognition. Centre Symmetric Local Binary Pattern (CSLBP) [13] is a notable improvement over LBP with respect to length and accuracy in region based image matching. Centre Symmetric Local Ternary Pattern (CSLTP) [14] is gradient based descriptor which performs well under controlled illumination changes. Semi Local Binary Pattern (SLBP) proposed in [26] computes region wise average values of grayscales and encodes the relationship of these average grayscales to generate the micropatterns. SLBP achieves better accuracy than LBP under noise and global illumination variations. Performance of CSLTP degrades under natural environment. Most recently a local descriptor namely; Local Directional Gradient Pattern (LDGP) [18] has been proposed which shows improvement over LDP[16], and LVP[17] with reduced length and comparable recognition accuracy. The performance of most of these descriptors under unconstrained environment is not comparable while applying on the facial image data set under constrained environment. The proposed descriptor has been designed to encode larger local region with minimum length to improve recognition and retrieval rates over constrained and unconstrained datasets.

The rest of the paper is organized as follows. Section 2 elaborates the major contribution as well as the motivation behind designing the proposed descriptor. Section 2 also defines the proposed descriptor. Section 3 defines all evaluation parameters used in the experiments. Experiments and comparative results are elaborated in section 4. Various conclusions are drawn on the proposed descriptor in section 5.

## 2. Proposed Cascaded Asymmetric Local Pattern

### 2.1 Motivation

Distinctive feature existing in a facial image are confined to the local region of the reference pixel. Mostly the distinctive information exists in the horizontal and vertical directions. The proposed descriptor encodes the asymmetry in the vertical direction as well as the asymmetry that exists in the symmetry in the horizontal direction. Local regions are confined to the neighboring pixels at a distance $R = 1$ in most of the descriptors and all these pixels at $R = 1$ are encoded by comparing them with reference pixel. The discriminating features, which are encoded using pixels at $R > 1$ increases the length of the micropattern beyond the acceptable limits [13]. Hence some of the descriptors ignore the pixels at lower distances while encoding the pixels at higher distances, which adversely affect the performance of the descriptor. The problems that exists in the existing descriptors namely; CSLBP, CSLTP, LBP, SLBP and so forth, have been addressed in the proposed descriptor. The proposed descriptor encodes larger variable size neighborhood of the reference pixel using minimum number of binary bits. Eight neighborhood pixels of the reference pixel of an image at a distance $R = 1$ are encoded with only 6-bits. Result analysis confer that the proposed descriptor effectively captures distinctive and useful relationships which significantly increases the accuracy of the proposed descriptor in terms of recognition rate as well as retrieval accuracies under constrained and unconstrained environment.

### 2.2 Major Contribution

The proposed descriptor encodes pixels at different radial distances with constant number of bits. At each of the radii eight neighbors are encoded with only 6 bits. The number of bits used by the proposed descriptor to encode the local neighborhood is very less as compared to LDP, LVP, SLBP and LBP. SLBP is a local descriptor that takes average values over $2 \times 2$ blocks to compute the LBP over these averages. The average taken over two by two blocks by SLBP removes meaningful information from the average image. The major drawback of CSLBP and CSLTP is that they ignore large number of pixels in encoding the descriptor. The higher order hand crafted descriptors namely; LDGP, LVP, LDP and so forth consider eight neighbors of the reference pixel. However, the length of these descriptors is very large. The proposed cascaded asymmetric local pattern (CALP) encodes 8 pixels in the local neighborhood of the reference pixel for $R = 1$, $R = 2$, $R = 3$ with 6 bits, 12 bits and 18 bits respectively. CALP computes the feature of length 64×3 bin for $R = 3$ which is notably less compared to the 256 bin features computed by LBP and SLBP. Table 1 shows the feature lengths of the descriptors, which clearly indicates that CALP effectively encodes large number of neighbors with optimal feature length.

### 2.3 Proposed Method

Most of the descriptors proposed for facial images tend to capture the relationships among the pixels around reference pixels. These relationships are encoded using binary values. These patterns are designed in such a way that they remain intact with changing pose, illumination, background and expressions. Proposed cascaded asymmetric local pattern (CALP) captures relationship amongst the pixels in horizontal and vertical directions to compute the asymmetry.

Local neighborhood of the encoding template is shown in Figure 1. The pixel being encoded with respect to the local neighborhood is called the reference pixel denoted by $I_{i,j}$. The template is the overlapping local block, which moves and encodes reference pixels. The local neighborhood of the reference pixel is divided into two pairs of three pixels. One pair consists of the pixels existing horizontally above and below the reference pixel and the other pair consists of the pixels existing vertically to the left and right of the reference pixels. These pairs are compared as shown in Figure 2 using (3) to generate the binary pattern shown in Figure 2 with black and white circles. Black circles denote 0 whereas 1 for the white circles.

$CALP_{i,j,D}^{H}$ is the decimal equivalent of the horizontal pattern computed using (1) by encoding the horizontal pixels above and below the reference pixel at a distance $D$. Encoding function shown in (3) has been used to generate the binary pattern. Similarly decimal equivalent of the vertical pattern $CALP_{i,j,D}^{H}$ is computed using (2).

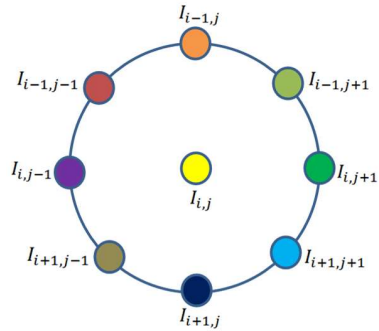

Figure 1. Template showing the local neighborhood of the reference pixel I$_{i,j}$.

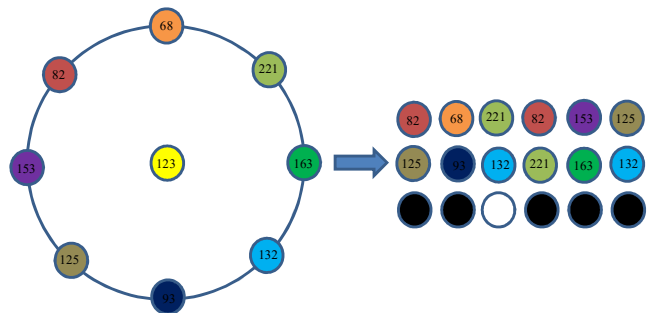

Figure 2. Example showing the encoding scheme of CALP.

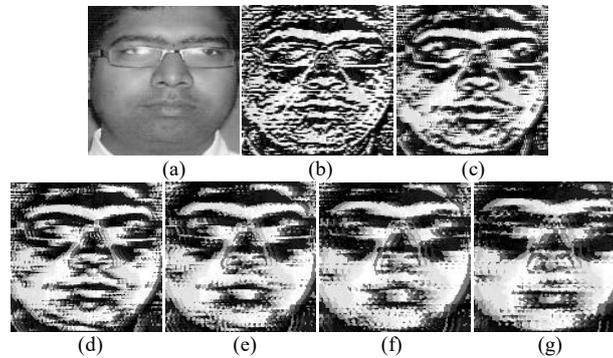

Figure 3. (a) Original image, (b) Feature image computed at D=1, (c) Feature image computed at D=2, (d) Feature image computed at D=3, (e) Feature image computed at D=4, (f) Feature image computed at D=5, (g) Feature image computed at D=6.

$$CALP_{i,j,D}^{H} = 2^5 \times C\big(I_{i-D,j-D}, I_{i+D,j-D}\big) + 2^4 \times C\big(I_{i-D,j}, I_{i+D,j}\big) + 2^3 \times C\big(I_{i-D,j+D}, I_{i+D,j+D}\big) \tag{1}$$

$$CALP_{i,j,D}^{V} = 2^2 \times C\big(I_{i-D,j-D}, I_{i-D,j+D}\big) + 2^1 \times C\big(I_{i,j-D}, I_{i,j+D}\big) + 2^0 \times C\big(I_{i+D,j-D}, I_{i+D,j+D}\big) \tag{2}$$

where $i = 1 + D .. M - D, j = 1 + D ... N - D$ and $C$ is the encoding function defined as

$$C(E,F) = \begin{cases} 0, & if E \le F \\ 1, & else \end{cases} \tag{3}$$

$CALP_{i,j,D}^{H}$ and $CALP_{i,j,D}^{V}$ are combined to compute the decimal equivalent of the 6b pattern generated from horizontal and vertical pixels in (4).

$$CALP_{i,j,D} = CALP_{i,j,D}^{H} + CALP_{i,j,D}^{V} \tag{4}$$

Finally histograms of $CALP_{i,j,D}$ denoted as $H_{CALP_{i,j,D}}$ are concatenated to form the $CALP$ feature vector as shown in (5).

$$CALP_R = \left\{ H_{CALP_{i,j,D}} \right\} | D = 1,2,3..R \tag{5}$$

where $i = 1 + D..M - D, j = 1 + D ...N - D$.

$\chi^2$ is the most commonly used similarity measure for the hand crafted descriptors [19]. Similarity measure $S_{\chi^2}(.,.)$ [35-36] is defined as

$$S_{\chi^2}(X,Y) = \frac{1}{2}\sum_{i=0}^{q} \frac{(x_i - y_i)^2}{(x_i + y_i)} \tag{6}$$

where $S_{\chi^2}(X,Y)$ is the $\chi^2$ distance computed on two vectors $X = (x_1, ..., x_q)$ and $Y = (y_1, ..., y_q)$. The probe set is classified using nearest neighbor classification. Chi-square ($\chi^2$) is used to compute the distance between probe and gallery images and class of the minimum distance gallery image is identified as the class of the probe image. The nearest neighbor classification has been used as it classifies the probe set with optimal computation cost [16] [35-36].

A sample of the local neighborhood and the corresponding encoding technique has been illustrated in Figure 2. A three bit pattern is computed using (3) by comparing the horizontal row of pixels in maroon, orange, and light green color with pixel in gray, deep blue, and light blue color respectively. Similarly, another three bits pattern is generated by comparing the vertical pixel in maroon, purple, and gray color with pixel in light green, green, and light blue color respectively. These three bit patterns are combined to form a six bit pattern as shown in Figure 2 with white and black pixels. Finally this six bit binary pattern is converted to equivalent decimal values to generate a feature image.

These feature images are shown in Figure 3(b-g) for the original image shown in Figure 3(a). Feature images shown in the figure clearly demonstrate that the information captured at larger distances has significant contribution in facial image identification.

The proposed descriptor is structurally different from state of the art descriptors such as SLBP (Year 2015), LDGP (Year 2017), LVP (year 2014), LDP (year 2010) etc. Major advantages of the proposed CALP descriptor are summarized as follows

1. SLBP divides the local region into 2×2 overlapping blocks and average values are computed over these 2×2 overlapping blocks and LBP is computed over these average values. As it takes the average over 2×2 blocks, significant amount of information is lost [35-36]. As opposed to SLBP, proposed CALP compares horizontal and vertical pixels across the reference pixel at different radius. The local neighborhood of SLBP is fixed whereas; the neighborhood of CALP can be increased by increasing the radius. It has been shown in the experimental results that with increasing size of the local neighborhood, CALP achieves improved recognition and retrieval accuracies. CALP (R=4) and SLBP have equal lengths. However, considerable gains shown by CALP (R=4) in recognition and retrieval accuracy across all databases as shown in section 4.

2. The higher order derivative space is encoded by LDGP using 6 bits whereas pixels of the original image are encoded in CALP. Only reference pixels are compared across higher order derivative space in LDGP, whereas horizontally and vertically opposite pixels of the original image are compared in CALP. The local neighborhood of LDGP is also fixed. Even though the length of LDGP is less than the length of CALP (R=2), CALP (R=3), CALP (R=4) the improved retrieval and recognition accuracy of CALP makes the contribution of this descriptor significant.

3. LDP and LVP are also derivative based descriptors which compare the pixels in the 3×3 neighborhood existing $0°, 45°, 90°,$ and $135°$ derivative space. Hence it can be said that LDP and LVP are structurally more close to LDGP. Similarly, LBP and CSLBP are structurally close, as they encode the neighborhood in the similar way. The structure of the neighborhood of CALP is similar to LBP and CSLBP. However, encoding scheme of CALP is different from LBP, CSLBP, LDGP, LDP and LVP. CALP captures dissimilarity in opposite regions of varying size of the local neighborhood. The feature lengths of LDP (32 bits) and LVP (32 bits) are more than the feature lengths of CALP (R=2), CALP (R=3), CALP (R=4), and CALP (R=5), which again confers the superiority of the proposed method.

The method of description of the facial image using separate horizontal and vertical block encoding of a neighborhood, which is different and effective as compared to the neighborhood used in LBP, LDP, LDGP, SLBP, CSLBP (diagonally opposite pixels) is new.

## 3. Performance Measures

Recognition and retrieval framework and corresponding evaluation parameters are used to show the effectiveness of CALP. Average Retrieval Precision (ARP) [28] [35-36] and Average Retrieval Rate (ARR) [28][35-36] are two of the performance measures used in retrieval framework. Precision is computed as

$$P_r(I_q, \lambda) = \frac{1}{\lambda} \sum_{i=1}^{|DS|} \Delta\big(\omega(I_q), \omega(I_i), \tau(I_q, I_i), \lambda\big) \mid I_i \neq I_q \qquad (7)$$

where $\lambda$ is the number of images retrieved, $I_q$ is the query image, $|DS|$ is the size of the dataset, $\omega(.)$ returns the class of an image and $\tau(I_q, I_i)$ is the rank of the $i^{th}$ image with respect to the query image $I_q$. Image rank is computed using similarity measure $S_{\chi^2}(.,.)$ between the $i^{th}$ image in the dataset and the query image. $\Delta(.)$ is a binary function defined as

$$\Delta\big(\omega(I_q), \omega(I_i), \tau(I_q, I_i), \lambda\big) = \begin{cases} 1, & \omega(I_q) = \omega(I_i) \ and \\ & \tau(I_q, I_i) \leq \lambda \\ 0, & else \end{cases} \qquad (8)$$

Average precision per class is computed as

$$AP(C_i, \lambda) = \frac{1}{|C_i|} \sum_{q=1}^{|C_i|} P_r(I_q, \lambda) \qquad (9)$$

where $C_i$ denotes the $i^{th}$ class in the dataset and $|C_i|$ denotes the number of images in the $i^{th}$ class. ARP is calculated over the entire dataset of $N_c$ distinct classes as

$$ARP(N_c) = \frac{1}{N_c} \sum_{i=1}^{N_c} AP(C_i, \lambda) \qquad (10)$$

Recall is defined as

$$R_e(I_q, C_i, \lambda) = \frac{1}{|C_i|} \sum_{i=1}^{|DS|} \Delta\big(\omega(I_q), \omega(I_i), \tau(I_q, I_i), \lambda\big) \mid I_i \neq I_q \qquad (11)$$

Average recall per class and ARR over the entire dataset are calculated as

$$AR_e(C_i) = \frac{1}{|C_i|} \sum_{q=1}^{|C_i|} R_e(I_q, |C_i|) \qquad (12)$$

$$ARR(N_c) = \frac{1}{N_c} \sum_{i=1}^{N_c} AR_e(C_i) \qquad (13)$$

F-Score [27] is another measure used to analyze the performance of the descriptors, which is defined as

$$F(N_c) = \frac{2 \times ARP(N_c) \times ARR(N_c)}{ARP(N_c) + ARR(N_c)} \qquad (14)$$

Table 1: Length of the descriptors

| Descriptor (year) | Length (bits) | Length (bins) |
|---|---|---|
| CSLBP (2009) | 4 | 16 |
| CSLTP (2010) | - | 9 |
| LDGP (2015) | 6 | 64 |
| LBP (1996) | 8 | 256 |
| SLBP (2015) | 8 | 256 |
| CALP (R=1) (proposed) | 6 | 64 |
| CALP (R=2) (proposed) | 6×2 | 64×2 |
| CALP (R=3) (proposed) | 6×3 | 64×3 |

Average normalized modified retrieval rank (ANMRR) defined in [20][27][35-36] is a performance measure used to identify which descriptor is actually retrieving more number of highly relevant images. Lower value of the ANMRR indicates that the descriptor retrieves images which are highly related to queried image or probe image. Similarly higher value indicates that the retrieved images are having higher rank (i.e. less relevant). The descriptor with highest F-Score should achieve least ANMRR value.

## 4. Performance Analysis and Discussion

Some of the most challenging facial image databases namely: Caltech-Face [21], CASIA-Face-V5-Cropped [22], Color FERET [23] [24], and LFW [25] are used to evaluate and compare the performance of the proposed descriptor with state of the art descriptors. It is evident from Table 1 that the length of the proposed descriptor at $R = 1$ is equal to the length of LDGP which is less than the length of LBP and SLBP. Similarly, length of CALP at $R = 6$ is comparable to the feature length of LBP and SLBP. The descriptors CSLBP and CSLTP have smaller lengths but the recognition and retrieval accuracies are less than the proposed descriptor. To compare the retrieval accuracy four evaluation parameters namely; ARP, ARR, F-Score and ANMRR are used. Two evaluation parameters namely; recognition rates and cumulative match characteristics (CMC) are used to compare the descriptors in recognition framework.

### 4.1 Performance analysis on Caltech-Face database

Caltech-Face database contains 450 frontal faces of 27 subjects with different backgrounds, expressions, and illuminations [21].

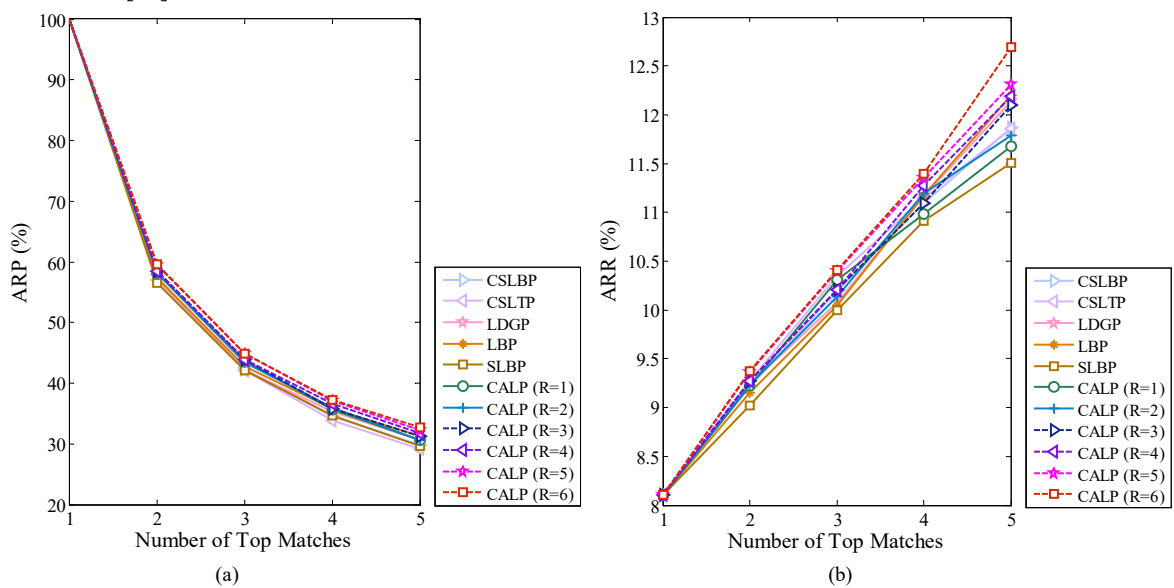

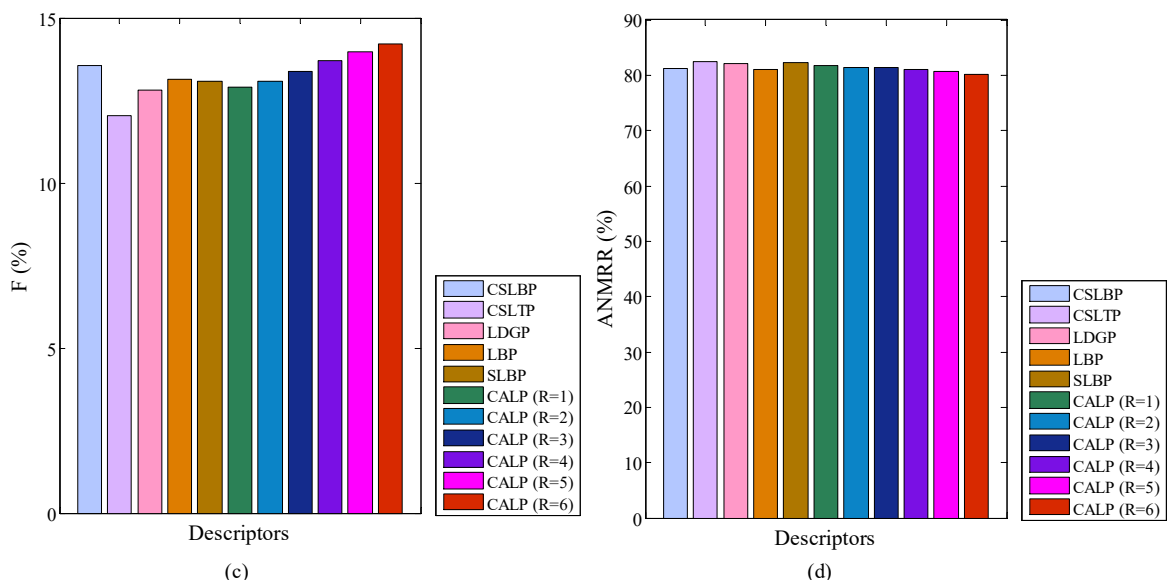

(c)                    (d)

Figure 4. (a) ARP, (b) ARR, (c) F-Score and (d) ANMRR. computed over Caltech-Face database.

Figure 4(a-d) shows the performance measures ARP, ARR, F-Score and ANMRR computed over Caltech-Face database. There is a 2%-3% improvement in ARP achieved by CALP over CSLTP. Improvement of 0.5%-1% has been achieved by CALP in ARR. Hence, the results show that the proposed descriptor CALP retrieves low rank (more relevant) images. CALP (R=6) achieves 1% higher F-score and 1% lower ANMRR than state of the art descriptors. Hence, it can be concluded that it retrieves more images having lower rank (images with low rank are closer to queried image).

## 4.2 Performance analysis on CASIA-Face-V5-Cropped database

"Portions of the research in this paper use the CASIA-FaceV5 collected by the Chinese Academy of Sciences' Institute of Automation (CASIA)" [22]. There are 500 classes in CASIA-Face-5.0 database where each class contains five color images. There are significant variation in intra-class images such as pose, illumination, imaging distance, eye-glasses, and expressions [22].

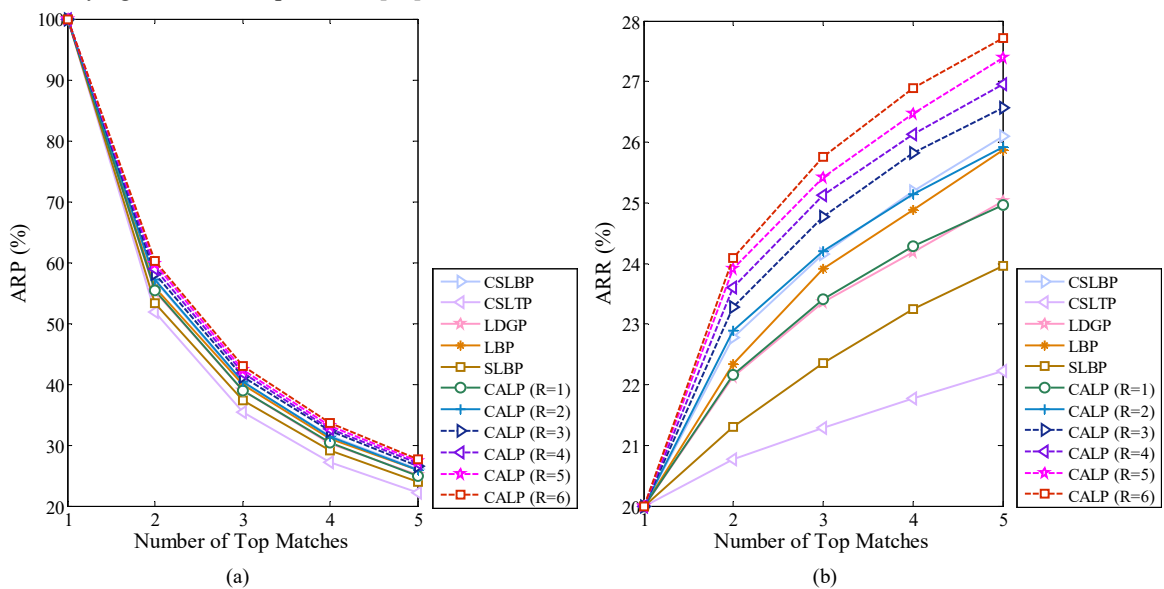

(a)                    (b)

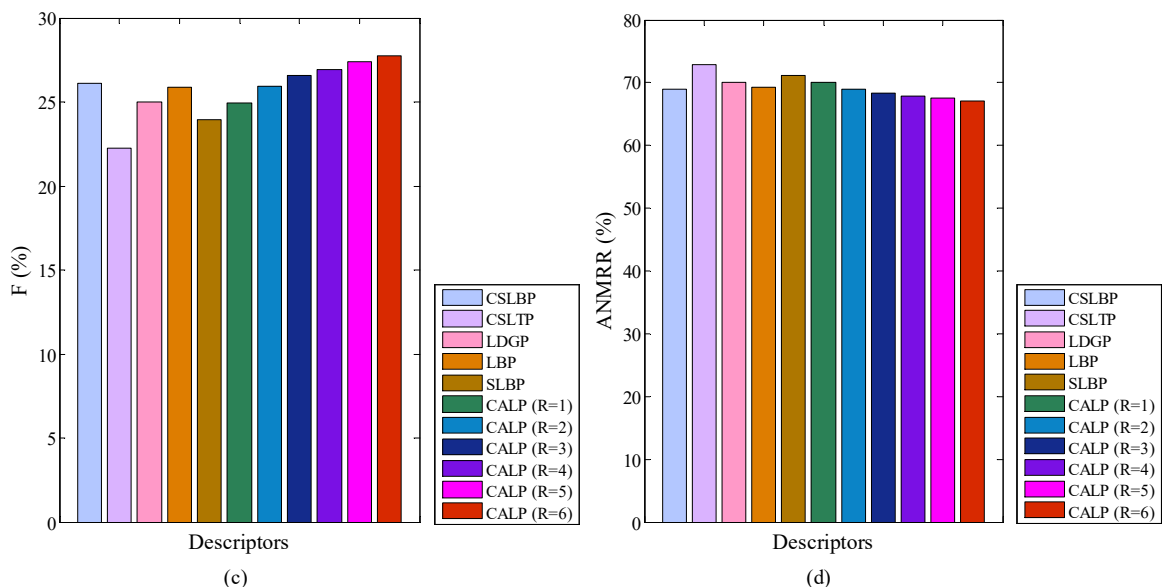

Figure 5. (a) ARP, (b) ARR, (c) F-Score and (d) ANMRR. computed over CASIA-Face database.

ARP and ARR of proposed descriptor CALP over CASIA-Face-5.0 database shown in Figure 5(a) and (b) confirms that approximately 2% better ARP and 2% better ARR over CSLBP, SLBP and LDGP have been achieved by CALP. The proposed descriptor also shows, 4% better ARP and 4% better ARR over SLBP. Higher F-score and lower ANMRR of CALP indicates that the most of the retrieved images are having lower/better rank or non-matching images are having higher rank.

### 4.3 Performance analysis on Color FERET database

"Portions of the research in this paper use the FERET database of facial images collected under the FERET program, sponsored by the DOD Counterdrug Technology Development Program Office". One of the most challenging unconstrained color facial image databases is FERET. There are 11,338 facial images of 994 individuals are in the database. All these images are taken under severe variations in pose and background. There are 13 different poses under which the images have been taken [23][24].

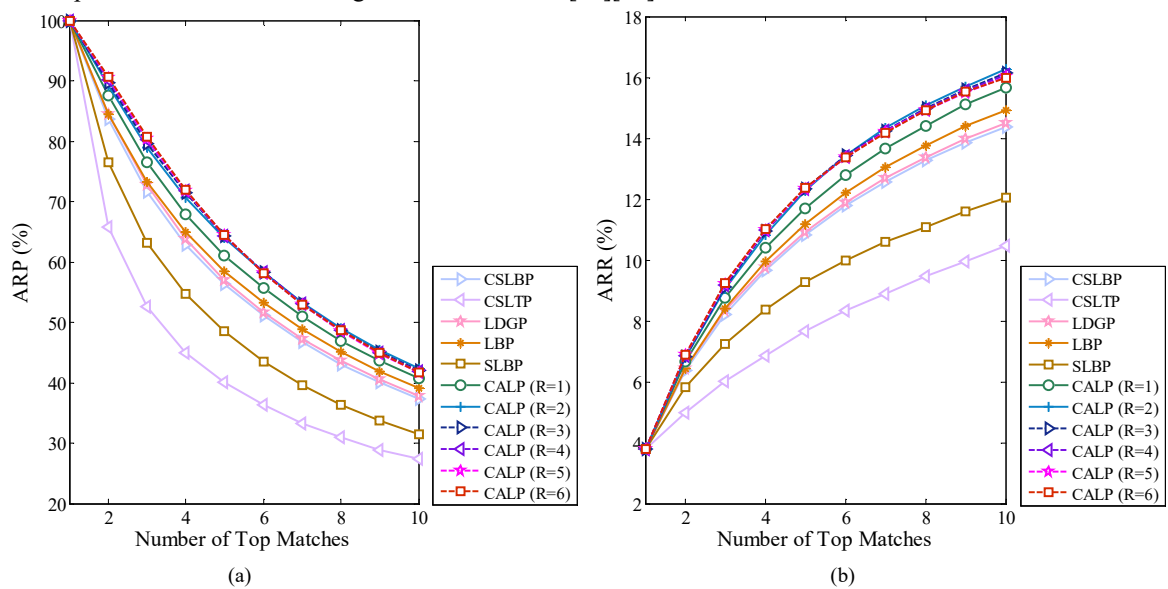

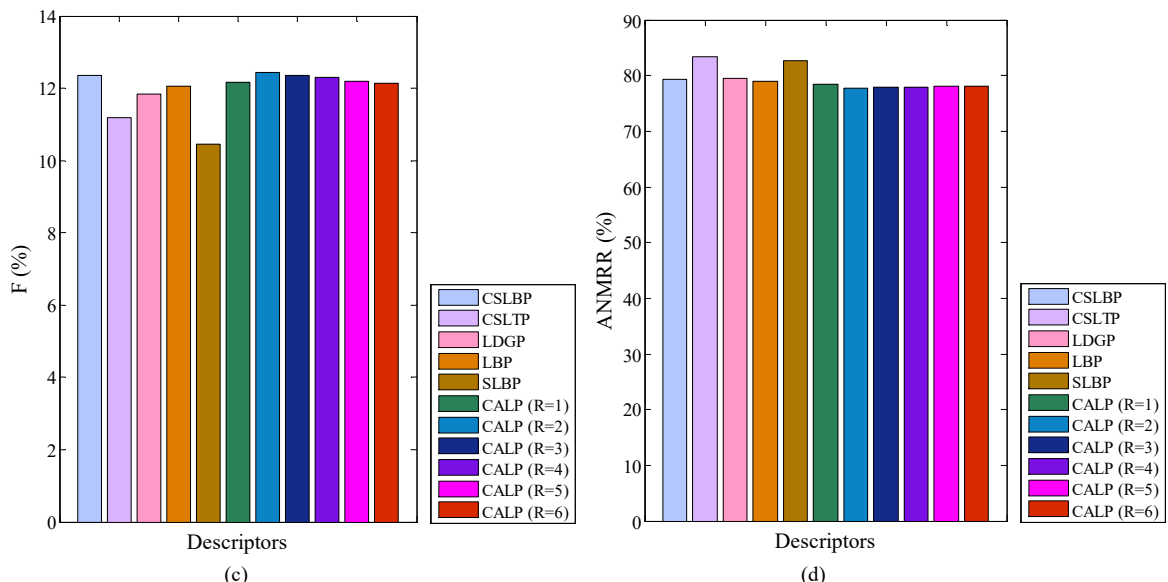

Figure 6. (a) ARP, (b) ARR, (c) F-Score and (d) ANMRR. computed over Color-FERET database.

The proposed descriptor is examined on color FERET database to show its capabilities against pose and expression variations. ARP and ARR values are shown in Figure 6(a-b) for a maximum of 10 retrieved images. The retrieval accuracy of LBP is closest to the retrieval accuracy of CALP. However, the accuracy of CALP is 2% more than the accuracy of LBP on FERET database. Even the length of LBP is comparable to the length of CALP.

F-Score achieved by CALP shown in Figure 6(c) is 0.5% higher than CSLBP, which is closest to CALP and ANMRR depicted in Figure 6(d) of the proposed descriptor is 1% less than CSLBP, LDGP etc. Lesser value of ANMRR indicates that CALP retrieves images, which have low rank and more relevant to probe image (images with low rank are closer to queried image).

### 4.4 Performance analysis on LFW database

LFW database has images of 5,749 individuals and there are 13,233 images in total [25]. The database contains two or more images of 1680 classes and the remaining classes have only one image each. The most challenging database under unconstrained environment is LFW and the proposed descriptor is also proposed for unconstrained environment. Hence the descriptor has been tasted on LFW.

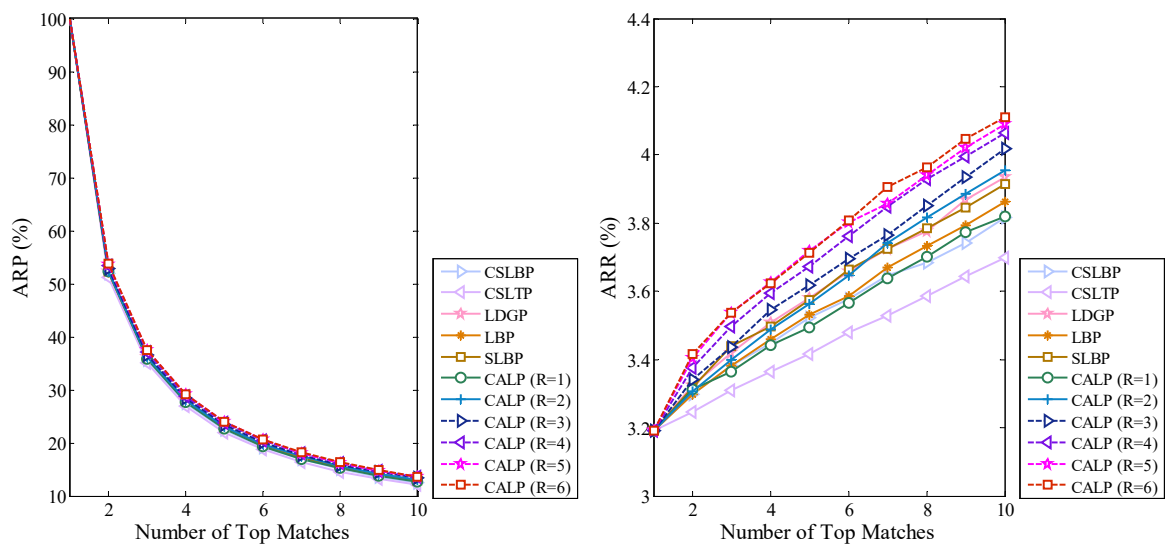

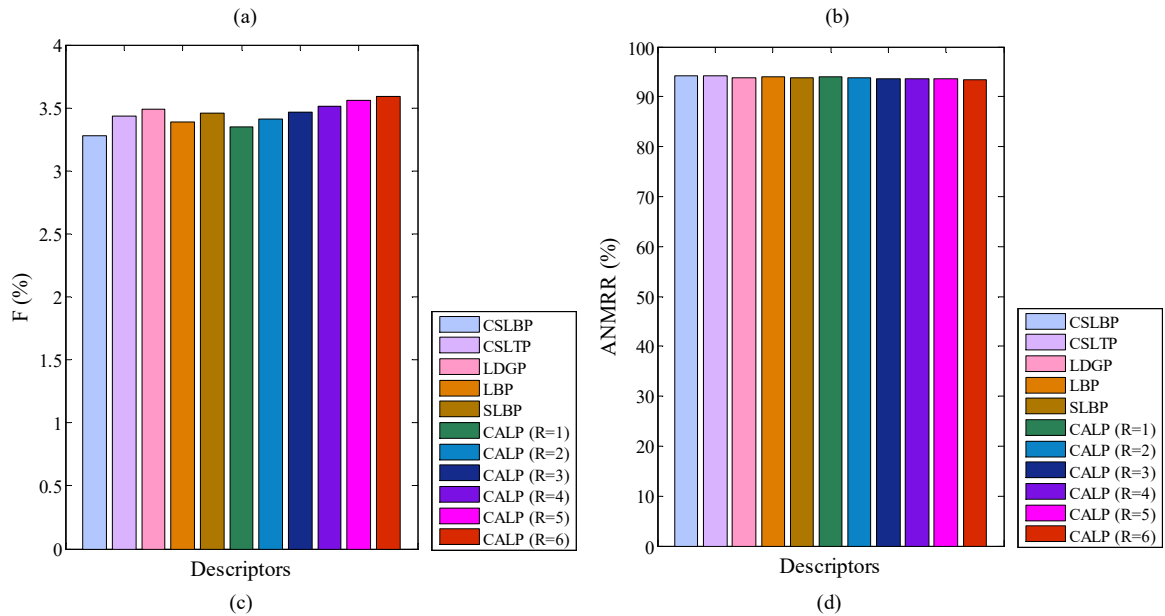

Figure 7. (a) ARP, (b) ARR, (c) F-Score and (d) ANMRR. computed over LFW database.

ARP and ARR over LFW are shown in Figure 7(a) and (b) respectively. Higher values of ARP and ARR of CALP descriptor prove that the retrieval performance is superior to the state of the art descriptors. Highest F-Score and Lowest possible value of ANMRR shown in Figure 7(c) and Figure 7(d) respectively for CALP indicate that it retrieves maximum number of most relevant images (low rank images are closer to queried image).

### 4.5 Performance analysis in recognition framework

A single probe image is compared with the rest of the images as gallery to compute the recognition rates. To compute the recognition rate one image is taken as probe image at a time from the database of $N$ images. The distance between probe and the rest of the $(N - 1)$ gallery images are arranged in ascending order. If the image with the smallest distance have the same class as the probe image then it is taken as a match, otherwise it is taken as a mismatch. $\chi^2$ distance measure is used to compute the distance between probe and the gallery images [35-36]. The Recognition Rate ($R_{reco}$) of a descriptor is computed as in (15).

$$R_{reco} = \left(\frac{number\ of\ matches}{N}\right) \times 100 \tag{15}$$

CMCs are computed for different descriptors over different databases and shown in Fig.8. Cumulative match is defined as the class match between the probe and gallery image with the rank less than or equal to the maximum retrieval rank specified.

The cumulative match score of the proposed descriptor is at least 10% more than the state of the art descriptors over Cattech_Face database. The proposed descriptor achieves 8% improvement over Color-FERET, 10% improvement over LFW and 10% improvement over CASIA-Face-V5-Cropped database. CALP achieves at least 8% improvement in cumulative recognition rate over its nearest state of the art counterpart. There is consistent gain in recognition accuracy of the proposed descriptor with increasing rank of the retrieved image.

The average recognition rate computed over variable size probe and gallery sets is shown in Fig.9. The entire database has been divided randomly into disjoint sets of probe and gallery. The probe sets made by extracting 20%, 30%, 40%, 50% and 60% images from the database. The remaining images are kept into the gallery set. [35-36]. Variable splits in probe and gallery are used to show the stable improvement in recognition accuracy even with decreasing size of gallery. The experiment is performed using 10-fold cross validation using different pairs of probe and gallery. The average recognition rate is computed over the recognition rates computed over 10 iterations of the experiment. This mechanism has been used to show that the average recognition accuracy of the proposed descriptor under random variations in the size of probe and gallery sets is maintained [35-36].

CALP shows significant improvement over other descriptors on Caltech-Face database. The descriptors CSLBP, CSLTP, LDGP, LBP, SLBP and CALP (R=2) achieve average recognition rates of 13.48%, 13.48%, 15.73%, 13.48%, 8.98%, and 20.22% respectively, for 20%-80% probe and gallery ratio. Average recognition rates of CSLBP, CSLTP, LDGP, LBP, SLBP and CALP (R=6) are 13.32%, 5.56%, 11.33%, 12.72%, 7.35%, and 19.08% respectively, over CASIA-Face-V5-Cropped database. The average recognition rates are 63.25%, 30.45%, 62.76%, 64.48%, 47.84%, and 74.84% of CSLBP, CSLTP, LDGP, LBP, SLBP and CALP (R=6), respectively, over Color-FERET database. Similarly average recognition rates of CSLBP, CSLTP, LDGP, LBP, SLBP and CALP (R=6) computed over LFW database 20%-80% ratio of the probe and gallery set are 6.77%, 5.61%, 8.92%, 7.27%, 7.93%, and 11.90% respectively.

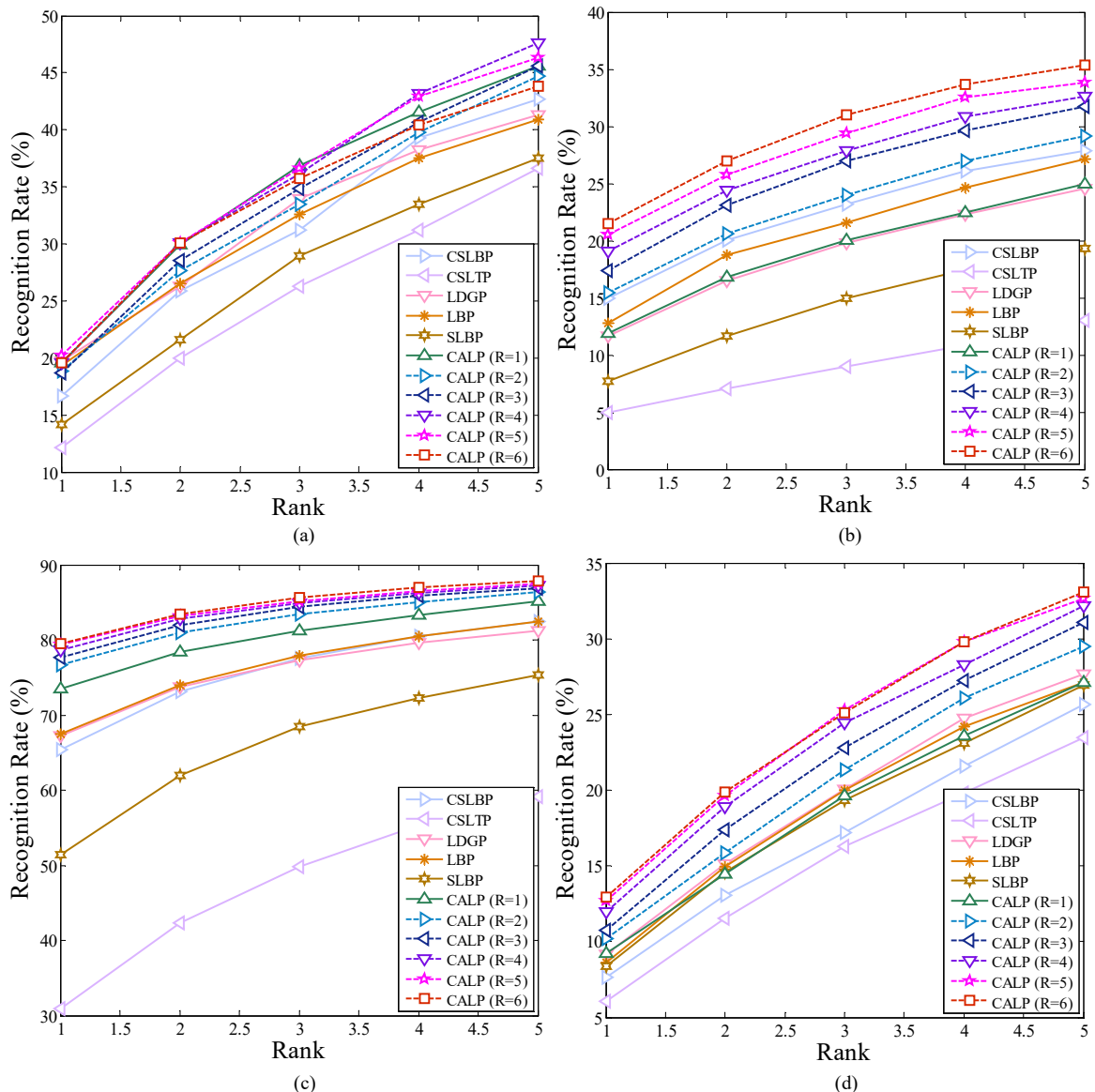

Figure 8. CMC for different descriptors and CALP on databases (a) Caltech-Face, (b) CASIA-Face-V5-Cropped, (c) Color-FERET, and (d) LFW.

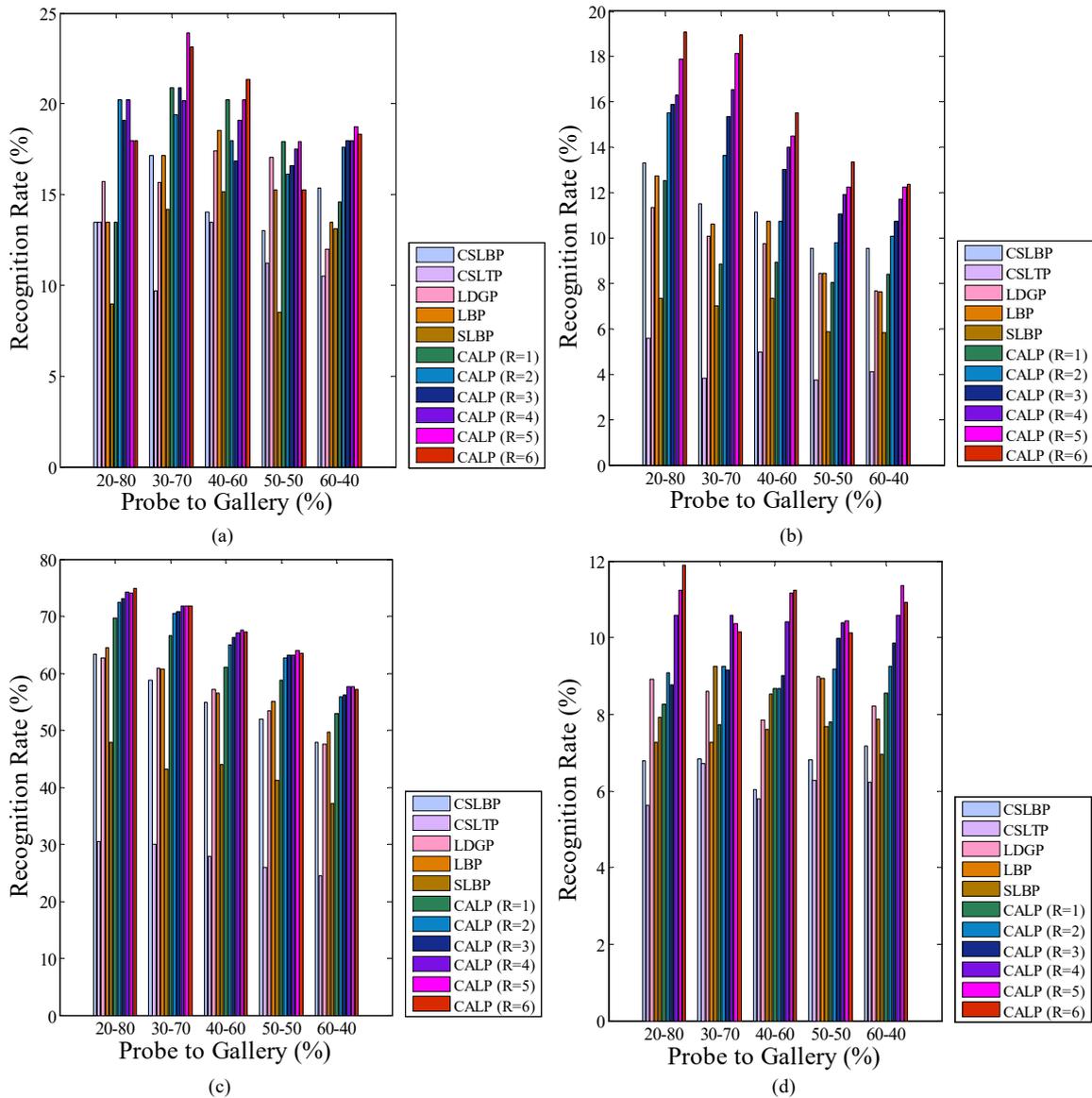

Figure 9. Comparative average recognition rates of different descriptors and CALP of 10-fold cross validation with different sized probe and gallery on databases (a) Caltech-Face, (b) CASIA-Face-V5-Cropped, (c) Color-FERET, and (d) LFW.

Extensive performance analysis over the most challenging databases shows that the proposed descriptor performs well under constrained as well as unconstrained databases.

## 4.6 Complexity Analysis

The computational complexity of the proposed descriptor has been computed by counting the number of primitive operations used to compute the micropatterns. In the local neighborhood the proposed descriptor requires six comparisons to compute a six bit pattern. The decimal conversions as shown in (1)-(2) require 4 additions and 6 multiplications which add up to 10 fundamental operations. One addition is required to compute $CALP_{i,j,D}$ from $CALP_{i,j,D}^{H}$ and $CALP_{i,j,D}^{V}$. Hence, the primitive operations required to compute the micropatterns for all the reference pixels over the facial image of size M × N at R = 1 are 17 × M × N. Therefore asymptotically, it can be concluded that the proposed descriptor has a complexity of $O(\text{M} \times \text{N})$.

## 5 Conclusion

The existing descriptors either encode only a limited number of pixels to reduce the feature length, which decreases the accuracy or encode the larger neighborhood with large number of pixels and increasing feature length. The descriptors encoding higher number of pixels in the local neighborhood employ such an algorithm to encode the neighborhood, which increases the feature length of the descriptor. The proposed descriptor encodes vertically and horizontally opposite rows with minimum number of binary bits to capture the horizontal and vertical asymmetry. Unlike CNNs the proposed descriptor is a hand-crafted descriptor and does not depend upon large training dataset and also applicable to various texture classification applications. Better retrieval as well as recognition accuracies of the proposed descriptor as shown in the experimental results illustrate that it outperforms the state of the art descriptors under constrained as well as unconstrained environments.